\documentclass{article} 
\usepackage{iclr2024_conference,times}

\usepackage{amsmath,amsfonts,bm}

\def\eqref#1{equation~\ref{#1}}

\def\1{\bm{1}}

\DeclareMathAlphabet{\mathsfit}{\encodingdefault}{\sfdefault}{m}{sl}
\SetMathAlphabet{\mathsfit}{bold}{\encodingdefault}{\sfdefault}{bx}{n}

\usepackage[hidelinks]{hyperref}

\usepackage{xcolor}
\hypersetup{
    colorlinks,
    linkcolor={red!80!black},
    citecolor={cyan},
    urlcolor={magenta}
}

\usepackage{url}

\usepackage{amsmath,bm}
\usepackage{graphicx}
\usepackage{microtype}
\usepackage{amsfonts}
\usepackage[noend]{algpseudocode}
\usepackage{algorithmicx,algorithm}
\usepackage{color}
\usepackage{etoolbox,lipsum}
\usepackage{float}
\usepackage{caption}
\usepackage{booktabs}
\usepackage{multirow}
\usepackage{array}
\usepackage{xspace}
\usepackage{enumerate}
\usepackage{nth}
\usepackage{cleveref}
\usepackage{bbm}
\usepackage{longtable}
\usepackage{CJKutf8}
\usepackage{tcolorbox}
\usepackage{xcolor}
\tcbuselibrary{breakable}

\title{TeleChat Technical Report}

\author{Antiquus S.~Hippocampus, Natalia Cerebro \& Amelie P. Amygdale \thanks{ Use footnote for providing further information
about author (webpage, alternative address)---\emph{not} for acknowledging
funding agencies.  Funding acknowledgements go at the end of the paper.} \\
Department of Computer Science\\
Cranberry-Lemon University\\
Pittsburgh, PA 15213, USA \\
\texttt{\{hippo,brain,jen\}@cs.cranberry-lemon.edu} \\
\And
Ji Q. Ren \& Yevgeny LeNet \\
Department of Computational Neuroscience \\
University of the Witwatersrand \\
Joburg, South Africa \\
\texttt{\{robot,net\}@wits.ac.za} \\
\AND
Coauthor \\
Affiliation \\
Address \\
\texttt{email}
}

\begin{document}

\maketitle

\begin{abstract}
In this technical report, we present \textbf{TeleChat}, a collection of large language models (LLMs) with parameters of 3 billion, 7 billion and 12 billion. It includes pretrained language models as well as fine-tuned chat models that is aligned with human preferences. TeleChat is initially pretrained on an extensive corpus containing a diverse collection of texts from both English and Chinese languages, including trillions of tokens. Subsequently, the model undergoes fine-tuning to align with human preferences, following a detailed methodology that we describe. We evaluate the performance of TeleChat on various tasks, including language understanding, mathematics, reasoning, code generation, and knowledge-based question answering. Our findings indicate that TeleChat achieves comparable performance to other open-source models of similar size across a wide range of public benchmarks. 
To support future research and applications utilizing LLMs, we release the fine-tuned model checkpoints of TeleChat's 7B and 12B variant, along with code and a portion of our pretraining data, to the public community.

\end{abstract}

\section{Introduction}
\label{intro}
The research community has witnessed substantial proliferation of open large language models (LLMs) that have emerged as valuable resources for study and as foundational models for developing chatbots and other applications. Following the introduction of ChatGPT ~\citep{chatgpt}, there have been thrilling advancements and applications of LLMs, but the majority of prominent LLMs, such as GPT-4 ~\cite{openai2023gpt4} and PaLM-2 ~\citep{anil2023palm}, are restrictive in their technological sharing. Few details about their models and training strategies are disclosed. This poses a challenge for developers and researchers who cannot access the complete model parameters, hindering a thorough examination or customization of these systems by the community. In contrast, a steady stream of openly accessible text-based LLMs has emerged, including OPT~\citep{zhang2022opt}, BLOOM ~\citep{bloom}, LLaMA ~\citep{llama}, LLAMA 2 ~\citep{llama2}, MosaicML's MPT ~\citep{mosiac}, Falcon ~\citep{penedo2023refinedweb}, etc. These models have provided researchers with valuable resources for further exploration and development, paving the way for extensive research in various domains, such as efficient fine-tuning techniques, longer prompt context utilization and retrieval augmented generation approaches. Furthermore, there exist various other LLMs that have been designed with a focus on Chinese language generation, including models such as Baichuan-2 ~\citep{yang2023baichuan}, Qwen ~\citep{bai2023qwen}, InternLM ~\citep{internlm} and SkyWork ~\citep{wei2023skywork}. While these models offer comprehensive details about their pretraining strategies, they often lack transparency in their instruction finetuning processes for chat models. This lack of transparency includes limited disclosure of the finetuning data composition, methods for concatenating multi-turn dialog data, and techniques employed to enhance conversational performance.

To encourage reproducibility of fine-tuned LLMs and foster responsible development of LLMs, we release TeleChat, a collection of chat models that have been fine-tuned using human alignment techniques including supervised fine-tuning and reinforcement learning. In particular, we provide a comprehensive explanation of our model architecture and the approach we used to extend TeleChat's context window to 96k in Section ~\ref{model}. Furthermore, in Section ~\ref{pretraining}, we delve into the specifics of our pretraining dataset and cleaning techniques. We then discuss alignment with human preferences in Section ~\ref{sft} and ~\ref{alignment}. Additionally, in Section ~\ref{experiment}, we conduct a thorough analysis of the model's performance on standard benchmark tasks. We also show our insights in Section ~\ref{graph} regarding mitigating hallucination with knowledge graph. Furthermore, we describe our parallel computing method in Section ~\ref{engineer}. Our contribution are listed as follows:
\begin{itemize}
\item[$\bullet$]We release TeleChat, a suite of pretrained and fine-tuned large language models (LLMs) with parameter sizes of 3 billion, 7 billion, and 12 billion. The foundation model of TeleChat undergoes pretraining on large corpus containing a diverse collection of English and Chinese texts, totaling trillions of tokens. Subsequently, TeleChat is fine-tuned to achieve state-of-the-art performance for conversational AI applications. The finetuned model of TeleChat's 7B and 12B variant is made public to the community. 

\item[$\bullet$] We present our comprehensive data cleaning workflow, which includes rule-based filtering and cleaning, deduplication at various levels (whole dataset, document, and paragraph), high-quality data selection, and data security processing. With our meticulous data cleaning approach, we ensure that TeleChat is pretrained on refined and reliable datasets. We also make available a portion of our high quality training corpus which includes 1TB of textual data.


\item[$\bullet$] We disclose a comprehensive description of our supervised fine-tuning methodology, an aspect that is frequently overlooked in reports of other publicly available models. Our methodology includes dataset blending, noisy embedding fine-tuning, and multi-stage long context training.

\item[$\bullet$] We provide our approach utilizing TeleChat for real-world applications, highlighting our methodology for mitigating hallucination through the use of knowledge graphs. Our objective is to assist the community in developing highly effective language models that can be applied across various practical scenarios.

\end{itemize}


\section{Model Design}
\label{model}
This section aims to provide an overview of our design methodology and shed light on the architecture of the TeleChat model. We begin by discussing the key components that form the model architecture. Subsequently, we elaborate on our approach to expanding the context window to 96k using interpolation and fine-tuning strategies.

\subsection{Model Architecture}
TeleChat is an autoregressive transformer model that employs a stack of transformer-decoder layers, whose architecture largely follows that of GPT-3 ~\citep{gpt3}. However, TeleChat deviates from the original transformer model in several notable ways, drawing inspiration from influential language models such as LLaMA ~\citep{llama2} and BLOOM ~\citep{bloom}. The key parameters of the architecture are summarized in Table ~\ref{architect1}.

\textbf{Rotary Position Embedding.}
We initially consider utilizing Attention with Linear Biases (ALiBi) to encode relative positional information, as proposed by ~\cite{press2022train}, due to its efficiency in implementation and extrapolation. However, the incompatibility of ALiBi with Flash Attention v2, which requires \texttt{attention\_bias} as an argument, led us to adopt Rotary Positional Embedding (RoPE, ~\cite{su2022roformer}) instead. Our decision was motivated by the successful implementation of RoPE in influential language models, such as LLaMA ~\citep{llama} ~\citep{llama2} and PaLM ~\citep{anil2023palm}, and its ability to extend context window lengths in recent studies ~\citep{chen2023extending}, ~\citep{rozière2023code}, ~\citep{peng2023yarn},~\citep{wei2023skywork}. By leveraging positional information through RoPE, we can efficiently encode absolute positions with explicit integration of relative position dependencies within the self-attention formulation. To further optimize computational efficiency and minimize memory usage, we implement Flash Attention v2 in the attention modules ~\citep{dao2022flashattention}, ~\citep{dao2023flashattention2}). Additionally, we choose to utilize \texttt{float32} precision for the inverse frequency matrix to prioritize model performance and achieve higher levels of accuracy.

\textbf{Normalizations.}
To ensure robust training, we incorporate an additional layer normalization step after the initial embedding layer for TeleChat's 3B variant, drawing inspiration from the methodology employed in BLOOM ~\citep{bloom}. However, we diverge from BLOOM by replacing conventional layer normalization with RMSNorm ~\citep{rmsnorm}, which has been shown to enhance the stability and performance of transformer models. Additionally, we adopted pre-normalization in each layer instead of post-normalization, a design choice that has been found to improve the training stability of transformer models.

\textbf{Activations}
We utilized the SwiGLU activation function ~\citep{shazeer2020glu}, a non-linear activation function that combines the strengths of Swish ~\citep{ramachandran2017searching} and Gated Linear Unit ~\citep{dauphin2017language}. SwiGLU has been shown to outperform other baseline activation functions, such as GeLU ~\citep{hendrycks2023gaussian}. We diminished the dimension of the feed-forward network to less than four times the hidden size, adhering to established conventions in prior research ~\citep{llama2} ~\citep{wei2023skywork}. In contrast to previous studies, our approach deviates from the convention of utilizing eight-thirds of the hidden size as the feed-forward network (FFN) dimension. Instead, we deliberately assign a specific dimension size to achieve the desired parameter size.


\begin{table}[htbp] 

\begin{center}
\begin{tabular}{ccccccc}
\toprule
     \textbf{Models}    & \textbf{layer num}     &  \textbf{attention heads}       &   \textbf{hidden size}   & \textbf{FFN hidden size} &  \textbf{vocab size} \\ 
\midrule
\midrule
 TeleChat-3B & 14 & 16 & 4096& 13312 & 82944 \\
 TeleChat-7B & 30 & 32 & 4096 & 12288 & 160256 \\
 TeleChat-12B & 38 & 32 & 5120 & 12288 & 160256 \\
\bottomrule
\end{tabular}
\end{center}
\vspace{-2mm}
\caption{Detailed model architecture parameters for TeleChat's 3B, 7B, and 12B models.}
\label{architect1}
\vspace{-1mm}
\end{table}

\subsection{Extending Context Window}
The input contexts of large language models (LLMs) can contain a substantial number of tokens in various scenarios, particularly when processing extensive inputs such as legal or scientific documents, database entries, and conversation histories, etc. ~\citep{schick2023toolformer}. As a result, it is crucial for LLMs to possess long-range capabilities and efficiently extrapolate to context lengths far beyond their initial pre-training limitations. 

In order to tackle the problem of losing high-frequency information during Position Interpolation (PI, ~\cite{chen2023extending}) on the RoPE embeddings, NTK-aware interpolation is proposed in ~\citep{ntk}. Instead of uniformly scaling each dimension of RoPE by a scaling factor, this approach redistributes the interpolation pressure across multiple dimensions by scaling high frequencies less and low frequencies more, which preserves high-frequency information ~\citep{peng2023yarn}. Furthermore, to address performance degradation resulting from fluctuations in context length during multiple forward-passes, we employ a Dynamic NTK-aware interpolation mechanism, in which the interpolation scaling factor is designed as a continuous variable, and is updated according to real-time context length.

To further enhance the long-context capabilities of TeleChat, we implement Multi-stage Long-context Training during the supervised finetuning phase and LogN-Scaling ~\citep{su2023improving} in the inference stage. Multi-stage Long-context Training periodically extends the context length during training, while LogN-Scaling adjusts the attention mechanism by rescaling the dot product in proportion to the context-to-training length ratio, ensuring the stability of attention entropy as the context length increases. The detailed description of Multi-stage Long-context Training can be found in section ~\ref{long}. Experimental results demonstrate that by employing these techniques, TeleChat successfully extends its context window to over 96k tokens.

\section{Pretraining Stage}
\label{pretraining}

During pretraining, we train the model from scratch using a substantial amount of data, which enables the model to not only gain a holistic comprehension of the world, but also develop specific skills such as mathematics, reasoning, and code generation. In this section, we introduce our data collection and cleaning method (Section ~\ref{data_collection} and ~\ref{data_process}), training details (Section ~\ref{pretrain_detail}), and tokenizer (Section ~\ref{tokenizer}).

\subsection{Data Collection}
\label{data_collection}

During the data collection stage, our paramount objective is to acquire a substantial and diverse dataset. This objective is accomplished through the collection of a vast amount of data from diverse sources, employing appropriate collection methods that guarantee a comprehensive representation of different perspectives.


\textbf{Data Sourcing.} TeleChat's pretraining corpus is curated from a wide range of data sources, constituting a comprehensive repository of knowledge. Our corpus contains both general-purpose and domain-specific data, ensuring a well-rounded and robust foundation. The general-purpose data comprises a vast range of sources, such as web pages, social platforms, encyclopedias, books, academic papers, code repositories, and more. In terms of domain-specific data, we gather corpus from twenty distinct sectors, including finance, construction, health and social work, aligning with national industry classifications\footnote{\url{https://www.stats.gov.cn/english/NewsEvents/200306/t20030619_25521.html}}. The specific textual formats we collect include financial report, bidding document, government notice, and various other document types.

\begin{table}[htbp] 
\begin{center}
\begin{tabular}{ccccccc}
\toprule
     \textbf{Datasets}  & \textbf{Percentage}\%    \\ 
\midrule
   \textbf{web page} &22 \\
   \textbf{books}  &11 \\
  \textbf{community QA} & 7 \\
  \textbf{social sharing} & 8\\
  \textbf{documents and reports } & 13 \\
  \textbf{paper} & 2 \\
  \textbf{code repository} & 12 \\
  \textbf{chat data} & 13 \\
  \textbf{others} & 12 \\
\midrule
\midrule
\textbf{Chinese} &45  \\
  \textbf{English} & 35 \\
  \textbf{Code}  & 11 \\
    \textbf{Math}  & 9 \\
  
\bottomrule
\end{tabular}
\end{center}
\vspace{-2mm}
\caption{The distribution of various categories of
TeleChat's pretraining data.}
\label{tab:pretrain}
\vspace{-1mm}
\end{table}

\textbf{Collection Method.} By leveraging the vast data repository accumulated by China Telecom over the past decades, our data collection process is simplified through the acquisition of a substantial volume of data from existing accumulations. In addition to textual data, our data collection procedure also contains the gathering of supplementary information, including timestamps, indicators of popularity (e.g. stars of GitHub repositories and numbers of likes/forwards of articles), and URLs. These supplementary details play a crucial role in the data filtering process discussed in Section ~\ref{data_process}. Furthermore, we diligently work towards enriching the data and mitigating biases. For example, we enrich our book data based on the Chinese Library Classification System\footnote{\url{https://en.wikipedia.org/wiki/Chinese_Library_Classification}}, and for social platform data, we employ a breadth-first search approach on social networks to encompass as many social groups as possible. Furthermore, we consistently gather and accumulate real-time data to ensure comprehensive coverage of the most up-to-date information.


During the data collection stage, we acquire diverse and extensive pre-training data on a petabyte scale, covering a wide range of domains. The distribution of our pretraining data is displayed in Table ~\ref{tab:pretrain}.

\subsection{Data Preprocessing}
\label{data_process}

During the actual collection process, various special situations may arise that necessitate specific subsequent handling. For example, the gathered data might include advertising, violent content, and private information that necessitates filtering. Additionally, there may be numerous duplicates and other redundant information. As a result, we devise a comprehensive data cleaning procedure to ensure the quality of our pretraining data. Our data clean procedure consists of rule-based filtering, deduplication, high-quality data selection, and data security filtering.  



\textbf{Rule-based Filtering.} Considering that TeleChat is primarily focused on Chinese and English, we remove data in other languages and non-text multimodal data. Simultaneously, heuristic rules are applied to clean the text efficiently and effectively. For instance, we filter out extremely short or low-information texts, discard texts with excessive or minimal punctuation, replace HTML tags with natural language, and automatically identify and standardize the text encoding format to \texttt{UTF-8}.


\textbf{Deduplication.} Performing global deduplication on a large amount of data is unacceptably slow, therefore we perform a hierarchical deduplication method that consist of \textit{URL deduplication}, \textit{Document-level Deduplication}, and \textit{Paragraph-level Deduplication}. First, we eliminate duplicate data from similar sources within groups using URL deduplication, which removes approximately half of the duplicate data. Next, we utilize a 128-bit SimHash algorithm to identify similarities in long texts, enabling Document-level Deduplication that removes duplicate articles, such as reposts on the internet. Finally, we employ Minhash and Jaccard similarity methods to perform Paragraph-level Deduplication, effectively filtering out a large number of homogeneous advertisements and heavily redundant texts. Notably, we use two different locally sensitive hash functions in the Document-level Deduplication and Paragraph-level Deduplication respectively, thereby achieving better deduplication results.



\textbf{High-quality Selection} We utilize a 5-gram Kneser-Ney model, as implemented in the KenLM library ~\citep{heafield-2011-kenlm}, to train on existing high-quality corpora and subsequently compute the perplexity of each paragraph. The lower the perplexity, the greater the similarity between the data and the high-quality corpora. It is important to note that the corpora used for training the model should be as unbiased as possible, however, it may still result in the erroneous discarding of relatively niche or highly specialized data. To address this, instead of simply discard texts with high perplexity, we split the data into three even parts: \textit{head}, \textit{middle}, and \textit{tail} based on the perplexity score, similar to ~\cite{ccnet}. The data in the \textit{head} part will be sampled more frequently, while the data in the \textit{tail} part will be sampled less during pretraining.


\textbf{Security Filtering.} To guarantee the security of our dataset, we utilize a multi-model classification approach that focuses on detecting and eliminating inappropriate, violent, and politically sensitive content. Our methodology prioritizes high recall and low precision, which enables us to identify a large number of negative instances that can be used for reinforcement learning. Additionally, we employ obfuscation techniques to safeguard personal privacy data, ensuring that sensitive information remains protected throughout the process.

\subsection{Training Details}
\label{pretrain_detail}

\textbf{Batch Generation.} To generate data batches, we employ a process of shuffling and concatenating the corpus obtained from the same source, ensuring consistency in the data. This approach deliberately avoids randomly concatenating data from different sources, thereby improving the model's ability to capture longer contexts. Furthermore, to align with the specified context lengths (e.g., 4096), the data is strategically truncated and concatenated with other data samples. By taking these steps, we can create batches of data that are not only diverse but also coherent, which is crucial for effective language modeling.

\textbf{Training Objectives.} The method utilized in the pretraining stage is known as autoregressive language modeling, which involves iteratively predicting the probability of the subsequent token in the sequence. We represent the joint probability of tokens in a text as:
\begin{equation}
    \begin{gathered}
        p(\mathbf{x}) = p(x_{1}, \cdot\cdot\cdot , x_{T} ) = \sum_{t=1}^{T} p(x_t|x_{<t})
    \end{gathered}
\end{equation}
Where $\mathbf{x}$ is a sequence of tokens, and we calculate the probability of each token $x_t$ based on the tokens that come before it, denoted as $x_{<t}$. The model is trained to optimize this probability across the entire training corpus.

\textbf{Optimizer.} We utilize the widely used Adam ~\citep{kingma2017adam} optimizer for pretraining optimization. We employ a cosine learning rate schedule, where the peak learning rate is specified for each model size. The learning rate gradually decays until it reaches a minimum learning rate of 10\% of the peak value. The hyperparameters are set as follows: $\beta_1 = 0.9$, $\beta_2 = 0.95$, and $\epsilon = 10^{-5}$. A weight decay of $10^{-4}$ is applied to all model parameters except for bias.

\textbf{Ramp-up Batch.} In order to enable the model to converge faster at the very beginning of pretraining, we employ a technique called ramp-up batch size, which involves starting with a small batch size and gradually increasing it linearly to the maximum batch size over a certain number of steps. 

\textbf{Precision}. The utilization of the float16 data type has been recognized as a possible factor contributing to numerical instabilities, leading to irreversible training divergences observed in large language models. This issue arises from the limited dynamic range offered by float16 ~\citep{zhang_bf16}. To ensure training stability, we pretrain all models using bfloat16 ~\citep{bfloat16} ~\citep{kalamkar2019study} ~\citep{bloom}, a data type that shares the same dynamic range as float32. In order to maintain a balance between performance and training stability, we employ bfloat16 mixed-precision training, as described by ~\citep{micikevicius2018mixed}. This approach involves performing precision-sensitive operations such as gradient accumulation, softmax, and weight updating with float32 precision, while carring out the remaining operations with bfloat16 precision.

The specific hyperparameters are presented in Table ~\ref{tab:param}.

\begin{table}[htbp] 
\begin{center}
\begin{tabular}{ccccccc}
\toprule
     \textbf{HyperParams}   & \textbf{TeleChat-3B}     & \textbf{TeleChat-7B}     & \textbf{TeleChat-12B}    \\ 
\midrule
\midrule
 \textbf{Peak lr} & 4e-4 & 3e-4 & 1.5e-4 \\
 \textbf{ramp-up batch size}  & 240/80/1,000,000 & 288/72/1,500,000 & 240/80/2,000,000 \\
 \textbf{batch size} &8M & 16M & 16M \\
 \textbf{warm up fraction} & 0.02 & 0.01 & 0.01 \\
 \textbf{clip-grad} & 1.0 & 1.0 & 1.0 \\
  \textbf{attention dropout} & 0.1 & 0.1 & 0.1\\
  \textbf{hidden dropout} & 0.1 & 0.1 & 0.1 \\
  \textbf{rmsnorm epsilon} & $1e^{-5}$ & $1e^{-5}$ & $1e^{-5}$ \\
  \textbf{\# training tokens} & 0.8T & 1.0T & 1.2T & \\
\bottomrule
\end{tabular}
\end{center}
\vspace{-2mm}
\caption{The hyperparameter details utilized during the pretraining stage of TeleChat's 3B, 7B, and 12B variants. The ramp-up batch size is expressed in the format of \textless start batch size \textgreater/\textless batch size increment\textgreater/\textless ramp-up samples\textgreater. For example, 240/80/1,000,000 indicates that the training begins with a batch size of 240 and increments by 80 for each time. The total ramp-up phase encompasses 1,000,000 samples.  }
\label{tab:param}
\vspace{-1mm}
\end{table}



\subsection{Tokenizer}
\label{tokenizer}
We utilize Hugging Face's tokenizers to implement the BBPE algorithm, training the tokenizer on a diverse dataset comprising Chinese, English, code, and mathematical data. This process results in a tokenizer with a vocabulary size of 160,130, which is subsequently padded to 160,256. Additionally, we use special tokens to differentiate dialogue roles and turns, and also incorporate specific designs to mitigate potential injection attacks.
\section{Supervised Fine-Tuning Stage}
\label{sft}
Large language models (LLMs) have demonstrated remarkable capabilities in various domains, such as reasoning ~\citep{wei2023chainofthought} ~\citep{yao2023tree}, coding ~\citep{chen2021evaluating}, ~\citep{Li_2022} and aligning general human intentions ~\citep{ouyang2022training}. Therefore, we employ supervised fine-tuning (SFT) stage after the pretraining stage to improve the model's ability to comprehend human behavior and effectively accomplish various real-world tasks. During the SFT stage, our model is exposed to various tasks using human-annotated prompts and feedback in a chat-style format. In this section, we provide detailed information about our data annotation method in Section~\ref{annotation}, followed by an in-depth discussion of our methodology and experimental details in Section ~\ref{method} and Section ~\ref{details}. Examples generated by TeleChat is shown in Appendix ~\ref{supp:examples}.

\subsection{Human Data Collection}
\label{annotation}




We have brought together a team of internal annotators and external contractors to carry out the manual data annotation process. Our annotators are all native Chinese speakers, boasting a range of academic backgrounds including Computer Science, Law, Chinese language and literature, and other related fields. This diversity enables them to excel in annotating expertise data with greater proficiency. Notably, a significant number of them hold bachelor or master degree from China's most esteemed 211 Project Universities \footnote{List of 211 Project Universities: \url{https://en.wikipedia.org/wiki/Project\_211}}. 

To ensure the high quality of our annotators and subsequently our data, we implement a rigorous training and selection process. This process starts with a comprehensive training session, providing our annotators with detailed instructional materials. They are then tasked with completing trial annotations, which are subsequently evaluated through a quality sorting mechanism. This allows us to assess the performance of each annotator and retain only those who have demonstrated exceptional proficiency in the assigned task. The data generated during these trials is not included in the final dataset to ensure data quality. 

We employ human annotators to label varied prompts and organize them into conversations, harnessing our annotation platform for efficient and high-quality annotations. During the labeling process, annotators are instructed to prioritize helpfulness, but for sensitive topics such as politics, violence, and pornography, they are instructed to prioritize safety and avoid harmful content. We work closely with labelers, providing them with clear instructions for each task and addressing their questions promptly. We continuously refine our instructions to ensure clarity and consistency, incorporating feedback from labelers to enhance the annotation process.

To further improve data quality, we implement a two-stage review process that includes checks by both reviewers and algorithm engineers. This process adheres to fundamental data requirements, such as fluency, helpfulness, truthfulness, and harmlessness, as well as domain-specific criteria, to guarantee data quality. Data samples that do not meet all predefined criteria according to a consensus among reviewers and algorithm engineers are excluded from the final dataset.

We collect over 100,000 supervised fine-tuning samples using the aforementioned annotation strategies and train our model accordingly. The statistics of the top 30 categories in our supervised-finetuning data is displayed in Appendix ~\ref{supp:sft}.

\subsection{Training Methodology}
\label{method}
In this section, we present a comprehensive explanation of our training approach during the supervised fine-tuning stage, an aspect that is frequently overlooked in reports of other open-sourced models. Our methodology contains the construction of SFT data samples (Section \ref{blend}), the usage of noisy embeddings for enhanced model performance in scenarios with limited training data (Section \ref{noisy}), and the implementation of multi-stage long-context training to expand TeleChat's context window to 96k tokens  (Section \ref{long}). By incorporating these techniques, we develop a chat model that offers valuable assistance and support to users.

\subsubsection{Data Organization}
\label{blend}
Our dataset spans various domains, such as General Q\&A, creative writing, reading comprehension, machine translation, code generation, math \& reasoning, and more. To ensure that each domain is represented appropriately, we assign respective resampling weights to each dataset based on their importance. Then, we sample single-round and multi-round conversations from each dataset using their corresponding resampling weights, thereby creating a balanced and diverse dataset that reflects the various domains. The sampled conversations are then shuffled and concatenated, followed by pre-padding them to a predetermined length (e.g., 4096 or 8192) to ensure consistent input length. We use special tokens \texttt{<\_user>}, \texttt{<\_bot>}, and \texttt{<\_end>} to denote the beginning of a question, the start of an answer, and the end of an answer respectively, thereby facilitating the model's comprehension of the conversational dynamics. To ensure diversity in the combination of data, the datasets are resampled and re-shuffled for each training epoch. We finetune the model in a supervised manner based on the instruction dataset.

\subsubsection{Noisy Embedding Fine Tuning}
\label{noisy}

In this section, we introduce our method for enhancing the answer quality of large language models (LLMs) through noisy embedding fine-tuning (\texttt{NEFTUNE}), inspired by the work of ~\cite{jain2023neftune}. Our approach involves introducing noise into the input embeddings of the LLM, which encourages the model to overfit less to the specifics of the instruction-tuning dataset. Instead, the model is more capable of providing answers that incorporate knowledge and behaviors of the pretrained base model. By doing so, we demonstrate that the conversational quality of the generated answers can be improved, and the model's ability to generalize to unseen tasks and data can be enhanced, especially when the training sample is limited.

Specifically, \texttt{NEFTune} modifies the input embeddings by adding a random noise vector to them. The noise is generated by sampling independent and identically distributed (i.i.d) uniform entries, each in the range $[-1, 1]$, and then scaling the entire noise vector by a factor of $\alpha/\sqrt{Ld}$, where $L$ is the sequence length, $d$ is the embedding dimension, and $\alpha$ is a tunable hyperparameter. This noise injection process simulates the variability and uncertainty present in real-world tasks, which helps the model to learn more robust and generalizable representations. 

However, we observe that while \texttt{NEFTune} can enhance the model's performance in scenarios with limited training data, its benefits diminish as the size of the training dataset increases. In fact, when an ample amount of training data is available, the impact of \texttt{NEFTune} becomes negligible. This is likely due to the model's reduced tendency to overfit on larger datasets. To investigate this further, we conduct experiments using TeleChat-7B fine-tuned models with and without the implementation of \texttt{NEFTune}. Our findings reveal that when \texttt{NEFTune} is applied, it achieves a 55\% win rate against its counterpart without \texttt{NEFTune}, as determined by human evaluators. However, when the model is trained on the entire dataset consisting of 4,000,000 samples, \texttt{NEFTune} loses its advantage, resulting in only a 48\% win rate against its counterpart without \texttt{NEFTune}. The effect of utilizing \texttt{NEFTUNE} is shown in Appendix ~\ref{supp:neftune}.





\subsubsection{Multi-stage Long-context Training.}
\label{long}
We utilize an innovative training approach involving a multi-stage method to enhance our model's capabilities in processing long-range context. During the supervised fine-tuning stage, we gradually increase the training length, enabling the model to activate and strengthen its ability to understand extensive dependencies while preserving its foundational skills. To achieve this, we periodically doubles the training length throughout the training process. This allows the model to encounter and learn from progressively longer contexts, leading to improved performance on tasks requiring a deep understanding of long-range contextual information. Specifically, we initiate the training with a sequence length of 8,192, building upon the foundation model trained on a sequence length of 4,096. At the 3/4 mark of the training procedure, we transit to a training sequence length of 16,384. Note that we employ the ntk-aware extrapolation method when working with sequence lengths of 8,192 and 16,384. This approach helps us mitigate the difficulties encountered during the transition, allowing for a smooth adjustment in the training sequence length for the model. Training details for TeleChat-7B's multi-stage long-context training is shown in Table ~\ref{tab:long}. Table~\ref{tab:long_Exp} presents the perplexity of TeleChat-7B on Wikipedia, demonstrating the effectiveness of incorporating NTK-aware extrapolation, attention scaling, and multi-stage long-context training.



\begin{table}[htbp] 

\begin{center}
\begin{tabular}{ccccccc}
\toprule
     \textbf{sequence length}    & \textbf{training steps}     &  \textbf{peak lr}       &   \textbf{batch size}   & \textbf{tensor parallel} &  \textbf{pipeline parallel} \\ 
\midrule
\midrule
 8,192 & 3,000 & 3e-5 & 8M & 2 & 4 \\
 16,384 & 1,000 & 4e-5 & 8M & 2 & 8 \\
\bottomrule
\end{tabular}
\end{center}
\vspace{-2mm}
\caption{Training details for TeleChat-7B's multi-stage long-context training. Note that training with a sequence length of 16,384 demands significantly more GPU memory compared to training with 8,192. As a result, it is necessary to increase the pipeline parallel size to 8, and requires 2 nodes to train. }

\label{tab:long}
\vspace{-1mm}
\end{table}

\begin{table}[htbp] \small
\begin{center}
\begin{tabular}{cccccccc}
\toprule
     \multirow{2}{*}{\textbf{Method}}    & \multicolumn{7}{c}{\textbf{sequence length}}   \\ 
     \cmidrule(lr){2-8} 
     & 2048 & 4096 & 8192 & 16384 & 32768 & 65536 & 98304 \\
\midrule
baseline & 4.8122 & 4.6562& 39.3099 & 98.3102 & 155.2708&487.3398 & 447.6295\\
NTK-aware (8k) & 4.8122 & 4.6562&5.1904 & 4.7155& 8.6351& 77.7478 & 79.9256\\
NTK-aware+logN (8k) & 4.8122& 4.6562& 5.1904& 4.0353& 4.1408& 9.4080& 7.9711\\
NTK-aware (16k) & 7.6916& 7.9900& 7.9580& 5.1217& 4.7932& 10.5444& 10.3614\\
NTK-aware+logN (16k) &7.6916 & 7.9900& 7.9580& 5.1217& 4.7195& 8.9751&7.6822 \\

\midrule
\bottomrule
\end{tabular}
\end{center}
\vspace{-2mm}
\caption{Our experiments with TeleChat-7B's long-context inferences illustrate the effectiveness of employing techniques such as NTK-aware extrapolation, attention scaling, and multi-stage long-context training. These approaches result in a significant reduction in perplexity as the context length increases and enable our model to achieve a low perplexity when extrapolating to 96K tokens. }

\label{tab:long_Exp}
\vspace{-1mm}
\end{table}

\subsection{Training Details}
\label{details}
During the supervised fine-tuning (SFT) stage, the model is initialized with the foundation model trained in the pretraining stage. Similarly to the pretraining phase, we employ next-token prediction as the training task. However, we introduce loss masks for user input questions to ensure that the loss is exclusively calculated for the output answer.

The model undergoes a total of 4,000 steps, with the first 3,000 steps involving training with a sequence length of 8,192, and the remaining 1,000 steps involving training with a sequence length of 16,384, as illustrated in section ~\ref{long}. In the training process, we utilize the same optimizer as in the pretraining stage, as described in section ~\ref{pretrain_detail}. The learning rate gradually increases over the first 10\% of steps until it reaches the peak learning rate. Afterwards, it decays using cosine decay to 10\% of the peak learning rate.

Moreover, to improve the stability of training large models, we apply global gradient norm clipping of $1.0$. To prevent overfitting, a dropout rate of $0.1$ is implemented, and a weight decay of $1e-5$ is applied to all model parameters except for bias. For efficient training, we utilize mixed precision training with dynamic loss scaling.


\section{Reinforcement Learning}
\label{alignment}

We also introduce reinforcement learning to align chat models with human preference, aiming to make model outputs consistent with safety and norms. 

\subsection{Reward Model}
When collecting prompts of reward dataset, a consensus is that high-quality and diverse prompts are conducive to the training stage of reinforcement learning.

We collect a large number of prompts, including data from both human annotation and internal user testing phases. The final prompt dataset consists of a total of 300 categories. To further get the high quality prompts, we use clustering and centroid selection to select representative prompts. All prompts are firstly convert to embeddings using bge-large-zh \footnote{\url{https://huggingface.co/BAAI/bge-large-zh-v1.5}}. Then we employ elbow clustering algorithms within each categories that aims to find the ideal number of clusters. The closest prompt to each cluster centroid will be selected. In addition, we randomly sampled the prompts in the cluster (except the closest prompt) to ensure the diversity of reward dataset, while the remain is used for reinforcement learning. The responses are collected from TeleChat models of different training stages and reasoning strategies, allowing sampling rich responses for annotation. 

Moreover, for improving the accuracy and reducing the difficulty of annotations, we simplify the task of ranking responses with human annotation. A straightforward classification task is introduced, where responses can be categorized under three distinct labels: good, medium, and bad. The basic criteria of this assessment includes but is not limited to safety, factuality, fluency, normality, etc. By evaluating the responses through these aspects, annotators can rank responses consistently. The responses between each pair of distinct labels under the same prompt can be combined with each other to form ranked pairs for subsequent training. 

\begin{table}[htbp] 

\begin{center}
\begin{tabular}{cccc}
\toprule
     \textbf{Type of data pairs}    & \textbf{good \& bad}     &  \textbf{medium \& bad}       &   \textbf{good \& medium} \\ 
\midrule
\textbf{Data Distribution} & 18.2\% & 21.1\% & 
 65.7\% \\
\textbf{Margin} & 1 & 2/3 & 1/3 \\
 \textbf{Test Accuracy} & 70.1\% & 66.0\% & 86.4\% \\
\bottomrule
\end{tabular}
\end{center}
\vspace{-2mm}
\caption{Training data distribution, adding margin and test accuracy of Reward Model on different type of data pairs.}

\label{tab:reward model}
\vspace{-1mm}
\end{table}

During the training stage, we use the same training objectives as LLaMA2 \cite{llama2}, adding margin in the loss function to teach the reward model to assign more difference scores to response pairs with more difference. The training data distribution, adding margin size and test accuracy of Reward Model on three types of data pairs are shown in Table \ref{tab:reward model}.

\subsection{Proximal Policy Optimization}

Proximal Policy Optimization (PPO) \cite{ppo2017} is widely used for LLM alignment and its mechanism is collaboratively working including four models: actor model, critic model, reference model and reward model. From the experience of \cite{yang2023baichuan} and \cite{bai2023qwen}, the critic model updates 50 steps firstly before actor model. The KL divergence coefficient is setting to 0.1 and apply a normalization process to the rewards, which accounts for the moving average. The learning rates for our actor and critic models are configured at $5 \times 10-6$ and $3 \times 10-6$ respectively through experiments. We get the chat model eventually after training for 400 steps.

\section{Engineering}
\label{engineer}

\subsection{Hardware}
TeleChat is trained on a total of 80 nodes, each having 8 Nvidia A100 Sxm 40GB GPUs. Each node is equipped with 2x Intel 6348 (28 Cores, 2.60 GHz) CPUs, 8x NVLink A100 GPUs, 512GB of RAM, and a 2GB cache RAID card. All nodes are interconnected using InfiniBand (IB) for networking. To enhance data transmission speed and mitigate bandwidth constraints, we employ NVIDIA's GPUDirect RDMA (GRDMA) and utilize the Scalable Hierarchical Aggregation and Reduction Protocol (SHARP). Training TeleChat took one month (including downtime).

\subsection{Parallel Computing}
TeleChat is trained using the Megatron-DeepSpeed framework ~\citep{megatrondeepspeed}, which is specifically designed for large-scale distributed training. By leveraging the capabilities of the Megatron-DeepSpeed framework, TeleChat benefits from 3D parallelism, which combines three complementary parallel approaches for distributed training.

\textbf{Tensor Parallelism} is a technique that partitions individual layers of a neural network model across multiple devices. In the training of TeleChat, the tensors for self-attention and feed-forward network module are partitioned along the row or column dimension, using a similar approach as mentioned in ~\cite{megatron_lm} . During the forward pass, the input tensor is distributed to each accelerator, which performs the computation simultaneously. After the forward pass, an all-reduce operation is performed to aggregate the results from all devices. This communication-intensive process is repeated four times per layer, twice for the forward pass and twice for the backward pass.

\textbf{Pipeline parallelism} is a technique used to parallelize the computation of a LLM by splitting its layers among multiple nodes. Each node represents one stage in the pipeline and receives inputs from the previous stage, performs computation, and sends the results to the next stage.

\textbf{Data Parallelism} involves replicating the model across multiple devices, dividing the global batch size among model replicas, and performing the training process in parallel, thereby leveraging the collective computational resources to accelerate the training process. After each training step, the model replicas synchronize to update their parameters. Increasing the global batch size enhances computational efficiency, but excessively large global batch sizes can lead to numerical instability during training, as discussed in references ~\citep{yuan} ~\citep{scaling_bz}. During the training process of TeleChat, we limit the global batch size to a maximum of 16M tokens in order to prevent numerical divergence.

To enhance the efficiency of our system, we implemente the Zero Redundancy Optimizer (ZeRO) ~\citep{rajbhandari2020zero} technique, which allows different processes to store only a fraction of the data required for each training step. Specifically, we utilize ZeRO stage 1, where only the optimizer states are partitioned in this manner. Additionally, to conserve memory on accelerators and accommodate larger models, we employe the strategy of recomputing activations during backward propagation, as described in ~\citep{korthikanti2022reducing}. 

By integrating these components, we scale our system to utilize hundreds of GPUs with extensive GPU utilization, achieving a peak performance of 180 TFLOPs using A100 GPUs, which accounts for 57.6\% of the theoretical peak performance of 312 TFLOPs.

\section{Experiment}
\label{experiment}

In this chapter, we evaluate the zero-shot and few-shot capabilities of TeleChat from various perspectives using standard benchmarks.
To fairly evaluate the performance of TeleChat, we select a list of models which have similar parameter sizes with TeleChat:
\begin{itemize}
\item[$\bullet$] LLaMA 2 (\cite{llama2}): LLaMA 2 is an upgrade of LLaMA, incorporating a larger amount of training data. LLaMA 2-Chat is fine-tuned on LLaMA 2, aligned with human preferences, enhancing the model's safety and usability.
\item[$\bullet$] InternLM-7B (\cite{internlm}): InternLM-7B is an open-sourced chat model. It utilizes trillions of high-quality data during the training process.
\item[$\bullet$] Baichuan 2 (\cite{yang2023baichuan}): Baichuan 2 is trained on 2.6 trillion tokens and has a significant improvements over Baichuan 1. In addition, it is optimized on solving math and code problems, with a impressive performance on medical and legal domain tasks.
\item[$\bullet$] ChatGLM 2-6B: ChatGLM2-6B is an open-source bilingual conversational model for both Chinese and English language. It is pre-trained on 1.4T tokens.
\item[$\bullet$] ChatGLM 3-6B : Based on ChatGLM 2-6B, ChatGLM 3-6B introduces more diverse set of training data and adopts a newly designed prompt format, which inherently supports complex scenarios such as Function Call, Code Interpreter, and Agent tasks.
\item[$\bullet$] Qwen (\cite{bai2023qwen}): Qwen is a language model developed by Alibaba. It has been trained on 3 trillions tokens of texts and codes. For chat models, Qwen has undergone RLHF to align with human preference. Furthermore, Qwen has received specialized reinforcement in areas such as code, mathematics, and agent functionalities.
\end{itemize}
\subsection{Examination Test Performance}
We evaluate TeleChat on multiple challenging examination test benchmarks. The questions in these datasets are also difficult for humans, requiring the model to possess extensive world knowledge and problem-solving capabilities to answer correctly. Therefore, these tests serve as a comprehensive measure of the model's abilities.
The detailed information of test benchmarks is as follows:
\begin{itemize}
\item[$\bullet$] MMLU (\cite{hendryckstest2021}): An English benchmark covering 57 tasks, which are mostly college level.
\item[$\bullet$] CMMLU: A Chinese benchmark to evaluate a LLM's knowledge and reasoning ability under Chinese scenarios.
\item[$\bullet$] C-Eval (\cite{huang2023ceval}): A comprehensive Chinese benchmark, containing more than 10 thousands questions and four difficulty levels.
\item[$\bullet$] GAOKAO-Bench (\cite{Zhang2023EvaluatingTP}): A Chinese evaluation benchmark utilizing Chinese college entrance examination questions (GAOKAO) to assess the language comprehension and logical reasoning abilities of LLMs.
\item[$\bullet$] AGIEVAL (\cite{zhong2023agieval}): A bilingual evaluation dataset, encompassing standardized test questions such as the Chinese National College Entrance Exam (GAOKAO), Law School Admission Test (LSAT), and Scholastic Assessment Test (SAT).
\end{itemize}

We have recorded the detailed experimental data in Table \ref{tab:open-benchmark}. To standardize the evaluation method, we employ the assessment technique provided by OpenCompass to obtain the results on most of the benchmarks. Specifically, MMLU, CMMLU and C-Eval were all conducted in a 5-shot setting, while the results for GAOKAO-Bench and AGIEVAL were achieved under a zero-shot method. The referenced model results all originate from the open leaderboard of OpenCompass.

We can observe that, compared to models of the same size, TeleChat exhibits superior performance. Particularly in terms of the results on the AGIEVAL and CMMLU datasets, TeleChat's performance surpasses that of other models of equivalent size (6-7B).
\begin{table}[!ht]
    \centering
    \begin{tabular}{cccccc}
    \toprule
        \multirow{2}{*}{Model} & MMLU & C-Eval & CMMLU & AGIEval & GAOKAO   \\
                       & (5-shot) & (5-shot) & (5-shot) & (zero-shot) & (zero-shot)\\
        \midrule
        LLaMA2-7B-chat & 46.2 & 31.9 & 31.5 & 28.5 & 16.1 \\ 
        LLaMA2-13B-chat & 54.6 & 36.2 & 38.7 & 32.3 & 18.6 \\ 
        ChatGLM2-6B-chat & 45.9 & 52.6 & 49.3 & 39 & 46.4 \\ 
        ChatGLM3-6B-chat & 51.9 & 53.8 & 54 & 38.9 & 49.3 \\ 
        InternLM-7B-chat & 52 & 54.1 & 52.6 & 43.7 & 45.8 \\ 
        Baichuan2-7B-chat & 52.8 & 55.6 & 54 & 35.3 & 39.7 \\ 
        Baichuan2-13B-chat & 57 & 56.7 & 58.4 & 40 & 51.4 \\ 
        Qwen-7B-chat & 56.6 & 59.3 & 59.5 & 41.3 & 63.3 \\ 
        Qwen-14B-chat & 66.4 & 71.7 & 70.0 & 47.3 & 76.5 \\ 
        \midrule
        TeleChat-7B-chat & 54.4 & 63.1 & 64.3 & 46.8 & 57.7 \\ \bottomrule
    \end{tabular}
    \caption{Results on benchmarks of Examination Test.}
    \label{tab:open-benchmark}
\end{table}
\subsection{Understanding Performance}
In addition to the exam test performance, we have tested TeleChat's comprehension abilities with traditional NLP tasks. We utilize three benchmarks: 
\begin{itemize}
    \item [$\bullet$] CSL (\cite{li2022csl}): A dataset containing 396k Chinese papers, which requires to checks the match between Chinese academic abstracts and their keywords.
    \item [$\bullet$] EPRSTMT (\cite{xu2021fewclue}): EPRSTMT is a sentiment analysis datasets based on comments on e-commerce websites.
    \item [$\bullet$] CHID (\cite{zheng-etal-2019-chid}): A reading comprehension benchmark, which requires the model to select the most appropriate idiom to fill in the blanks within the text.
\end{itemize}
The results are shown in Table ~\ref{tab:open-benchmark-other}. TeleChat-7B-chat outperforms the baseline models on the CSL and CHID datasets, which indicates that TeleChat has excellent comprehension capabilities. In practical applications, this kind of traditional NLP task still has a great effect, so it is reasonable to believe that our model can be well applied in application.

\begin{table}[!ht]
    \centering
    \begin{tabular}{ccccccc}
    \toprule
        \multirow{2}{*}{Model} & CSL & CHID & EPRSTMT & GSM8K & MATH & HumanEval \\ 
        & (zero-shot) & (zero-shot) & (zero-shot) & (4-shot) & (4-shot) & (zero-shot) \\ 
        \midrule
        LLaMA2-7B-chat & 58.8 & 44.1 & 57.5 & 26.3 & 3.9 & 12.2 \\ 
        LLaMA2-13B-chat & 61.2 & 48 & 59.4 & 29.6 & 5.0 & 18.9 \\
        ChatGLM2-6B-chat & 61.2 & 57.9 & 71.2 & 28.8 & 6.5 & 11 \\ 
        ChatGLM3-6B-chat & 65.6 & 63.4 & 85 & 56.7 & 18.7 & 61 \\ 
        InternLM-7B-chat & 70 & 79.7 & 88.8 & 34.6 & 5.6 & 12.8 \\ 
        Baichuan2-7B-chat & 60 & 75.2 & 87.5 & 32.8 & 6 & 13.4 \\ 
        Baichuan2-13B-chat & 63.1 & 78.2 & 87.5 & 55.3 & 8.6 & 17.7 \\ 
        Qwen-7B-chat & 63.1 & 72.3 & 88.8 & 52.5 & 10.3 & 26.2 \\ 
        Qwen-14B-chat & 55.6 & 72.3 & 91.2 & 61.0 & 26.8 & 36.6 \\ 
        \midrule
        TeleChat-7B-chat & 66.81 & 88.0 & 87.5 & 36.7 & 10.3 & 14.6 \\ 
        \bottomrule
    \end{tabular}
\caption{Results on benchmarks of Understanding, Reasoning and Coding Performance.}
    \label{tab:open-benchmark-other}
\end{table}

\subsection{Reasoning and Coding Performance}
To test the reasoning and coding capabilities of the model, we used the following three datasets:
\begin{itemize}
\item [$\bullet$]GSM8K (\cite{cobbe2021gsm8k}): GSM8K is a dataset of 8.5K high-quality, linguistically diverse, human-written elementary math problems.
\item [$\bullet$]Math (\cite{hendrycksmath2021}): A dataset containing 12.5K challenging competition math problems.
\item [$\bullet$]HumanEval (\cite{chen2021codex}): A code capability test dataset provided by OpenAI, which consists of 164 programming questions that measure the correctness of code 
\end{itemize}
According to Table ~\ref{tab:open-benchmark-other}, TeleChat-7B-chat's reasoning and coding performance among models with 6B-7B parameters is second only to chatglm3-6b and qwen-7b-chat, which is significantly better than other models of the same size. However, the relatively low performance of TeleChat-7B-chat compared to the 13B-14B model may be related to the significant effect of model parameter size on math and coding ability.

\section{Alleviating Hallucination with Knowledge Graph}
\label{graph}

Hallucination problems are frequently observed in LLMs, where there is a tendency to generate text that appears coherent and meaningful but lacks real-world existence. This can cause confusion and misunderstandings for users who rely on such information for decision-making. Hallucination problems in LLMs can be classified into two types: deviation from established world knowledge, and lack of coherence with the source context. In this study, we address the first type of hallucinations by utilizing structured information representation provided by Knowledge Graphs (KG).


\begin{figure}[t] 

    \begin{center}
        \includegraphics[width=0.9\linewidth]{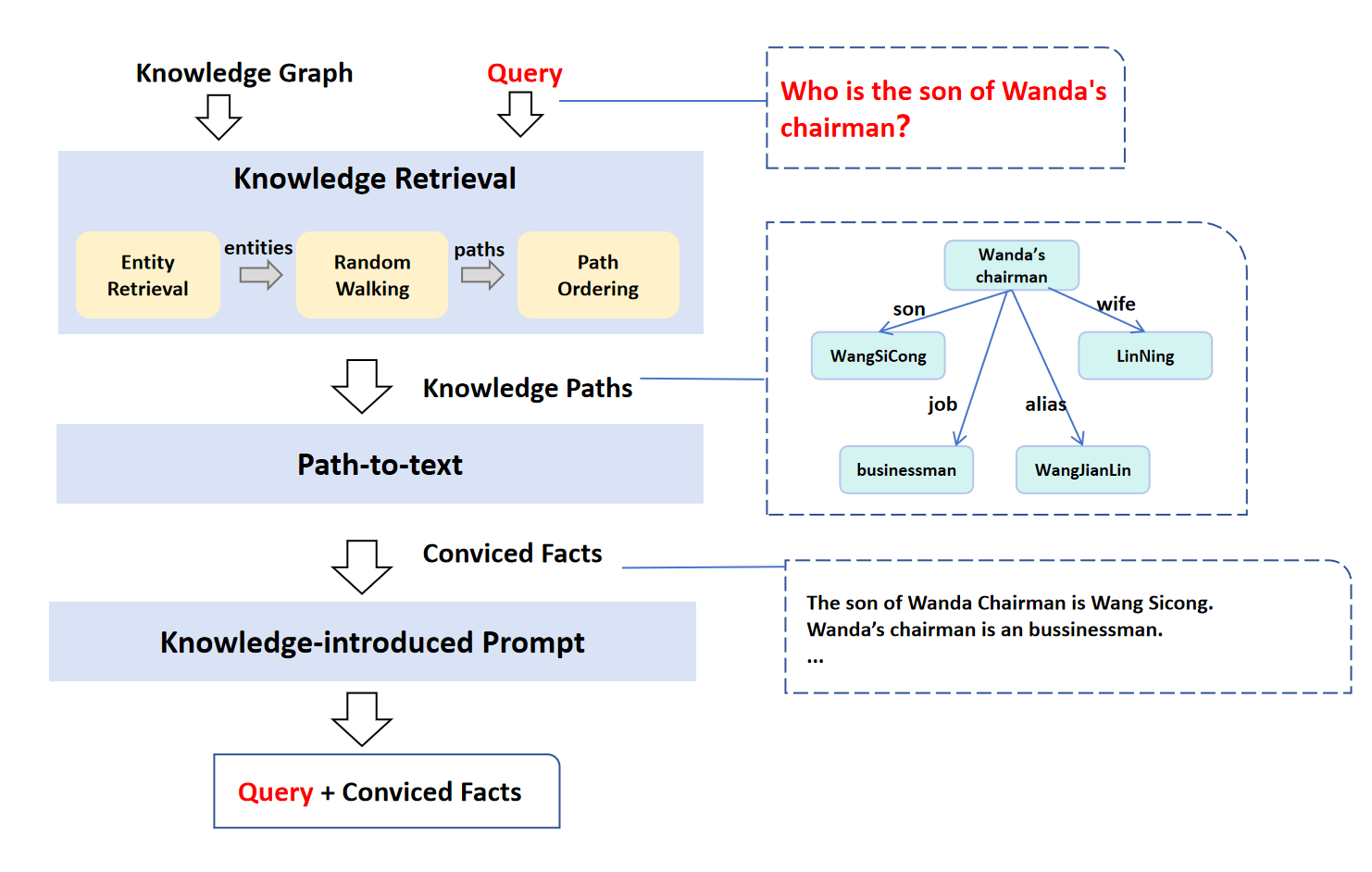}
    \end{center}

    \vspace{-2mm}
    \caption{\textbf{The overall process of introducing knowledge into prompts.}}
    \label{fig:kg-intro}
    \vspace{-2mm}
\end{figure}

The overall operational process of introducing knowledge into prompts is shown in Figure \ref{fig:kg-intro}. When a query comes, candidate entities are firstly retrieved based on n-gram similarity with query. Subsequently, a random walk of n steps is conducted within the graph, starting from these candidate entities. Finally, all paths obtained through the random walk are sorted based on their relevance to the user's query. The top-k paths are then returned as the final result of the knowledge graph retrieval process. By combining this retrieved knowledge with a prompt, the large language model can process the augmented query, taking into consideration the background knowledge provided by the knowledge graph. This approach helps mitigate the risk of hallucinations, as the model gains a more accurate understanding of the real-world relationships and entities associated with the given source content.


We evaluated the TeleChat's ability to answer factual questions in the China Conference on Knowledge Graph and Semantic Computing (CCKS) 2020 Knowledge Graph based Q\&A task\footnote{\url{https://sigkg.cn/ccks2020/?page_id=69}}. Without the introduction of the knowledge graph, the accuracy of TeleChat on this task is recorded at $0.19$. However, after incorporating the relevant knowledge by adding the top 10 relevant paths from the knowledge graph, the accuracy significantly improves to 0.69. This demonstrates the effectiveness of integrating the knowledge graph in enhancing the TeleChat's ability to provide accurate answers to factual questions.

\section{Related Work}

Language modeling has been a central problem in natural language understanding, which models the probability distribution of the occurrence of text sequences. Since the emergence of the Transformer structure (\cite{vaswani2017attention}), which is based entirely on the attention mechanism, it is possible to train millions of parameters or even billions of parameters utilizing parallel computation on  large number of matrix computations.
As language models based on the Transformer structure continue to emerge, they are gradually replacing traditional language models. By using more training corpus on a larger scale model, these new language models have achieved breakthroughs in effectiveness.
During that period, BERT (\cite{devlin2018bert}), GPT-2 (\cite{radford2019language}) and T5 (\cite{raffel2020exploring}) became the most representative language models, which achieved remarkable success on traditional natural language processing tasks. Then, OpenAI introduced the GPT-3 (\cite{brown2020language}) model with an astonishing 175 billion parameters. With the introduction of models such as PaLM (\cite{anil2023palm}) and Bloom (\cite{bloom}), the model size continues to grow following the law of scaling (\cite{kaplan2020scaling}). The proposal of chain of thought (CoT) (\cite{wei2023chainofthought}) has highlighted the great potential of large models and attracted widespread attention.

In 2022, OpenAI launched ChatGPT (\cite{chatgpt}), which once again broke people's inherent perception of artificial intelligence. It is equipped with powerful capabilities to effectively assist humans in accomplishing various tasks.
After that, OpenAI released GPT-4 (\cite{openai2023gpt4}), a model with more powerful language comprehension and even achieved scores above the human average in some college exams. 
Compared to previous large models, ChatGPT and GPT-4 have been optimized in the process of alignment with human preferences using so that they generate results that are more in line with human expectations and needs (\cite{ouyang2022training}). This optimization ensures that the models are better able to understand human intent and thus provide more accurate, safe, and useful responses.
Nevertheless, OpenAI has not open-sourced their model weights, placing certain constraints on developers in their application.
Fortunately, Facebook's release of LLama (\cite{llama}) allows the open-source community to further develop based on this model.
In practical development and application, due to limitations in inference speed and GPU memory size, models with 3 billion to 20 billion parameters are considered to be the most cost-effective. Therefore, large Chinese language models often take the initiative to experiment with models of this scale, and notable examples such as Baichuan (\cite{yang2023baichuan}), QWen (\cite{bai2023qwen}), and ChatGLM (\cite{zeng2022glm}) have all demonstrated commendable performance. 
To leverage the powerful capabilities of language models for addressing specific problems, the prevailing approach involves utilizing the decision planning ability of large-scale models as agents to connect multiple tools, which has become the current mainstream direction (\cite{schick2023toolformer,qin2023tool,zeng2023agenttuning}). 
In addition, to address the hallucination phenomenom during application processes, the integration of large-scale models with knowledge graphs has emerged as a highly acclaimed cutting-edge approach (\cite{pan2023unifying}).

\section*{Acknowledgement}
We would like to extend our gratitude to the following members of our team for their contributions in coordinating data collection - Jianzheng Song, Linlin Miao, Yanwu Zhao, Zhu Yuan, and many others. Furthermore, we would like to acknowledge the support and invaluable insights provided by the following participants - Chunping Jiang, Haidong Hu, Jiaxi Ma, Kaili Wang, Xinzhe Zhou, Yi Yao, Yuwei Jiang, Yuxiao Huang, Zhoubao Wang, Zhihua Duan, and many others. Please note that the individuals are listed alphabetically by their first names, and their order does not indicate the ordering of their contributions.

\bibliography{iclr2024_conference}
\bibliographystyle{iclr2024_conference}

\clearpage

\appendix


\section{Details of Supervised Finetuning Data}
\label{supp:sft}
\begin{figure}[ht]
    \vspace{-6mm}
    \begin{center}
        \includegraphics[width=1.0\linewidth]{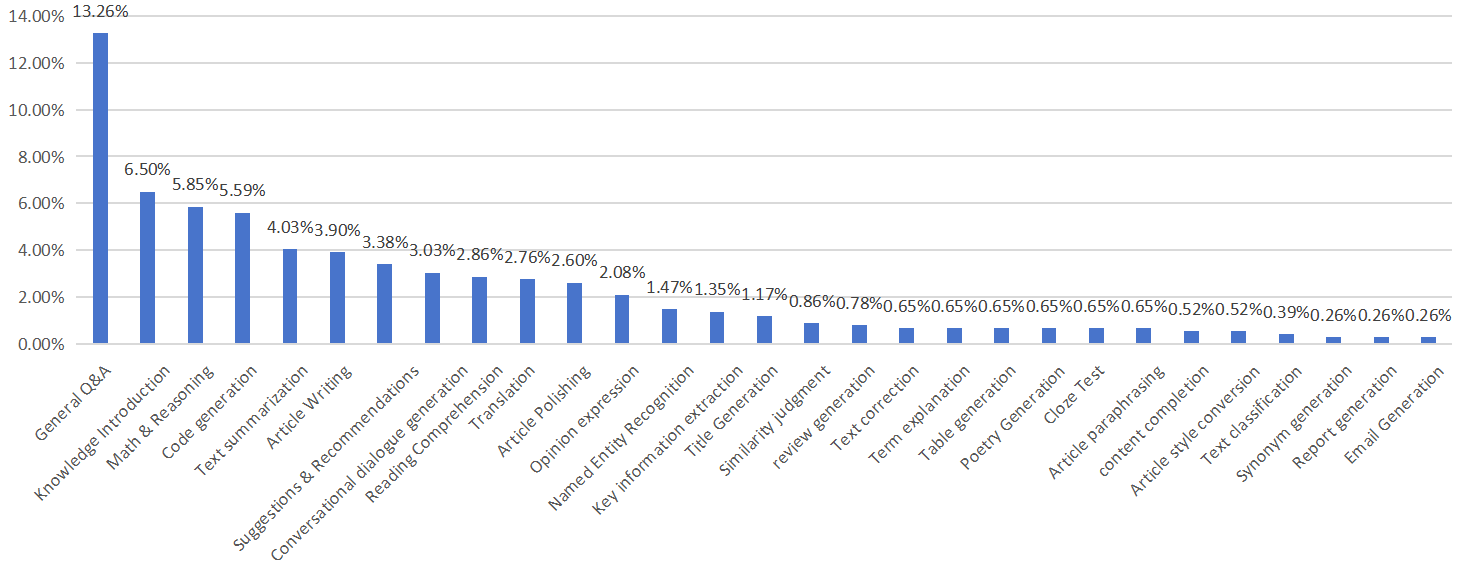}
    \end{center}
    \caption{Illustration of the Top 30 categories in our SFT data.}
    \label{fig:sft}
    \vspace{-3mm}
\end{figure}

This appendix presents the statistics of the top 30 categories in our supervised-finetuning data, based on percentage of sample numbers. While our dataset contains over 100 categories in total, we only highlight the top 30 in Figure~\ref{fig:sft} for clarity and brevity. In order to expand our dataset, we annotate custom output template for each identified category and expanded the sample set through manual augmentation accordingly.
\\
\\

\section{Examples generated by TeleChat}
\label{supp:examples}
\subsection{Office assistance}
\begin{tcolorbox}[colback=gray!10!white,colframe=gray!50!black,breakable]
\begin{CJK}{UTF8}{gbsn}
\noindent\textbf{Instruction:} 

给出以下信息，请你写一篇完善的工作计划。\\
1. 分析市场竞争情况，包括主要竞争对手和他们的产品特点。\\
2. 制定产品开发路线图，包括关键里程碑和时间表。\\
3. 确定资源需求，包括人力、资金和技术支持。\\
4. 开展市场调研，以了解潜在客户的需求和反馈。\\
5. 制定市场推广策略，包括定价策略、促销活动和渠道选择。\\
请输出详细的项目计划。\\(Given the following information, write a well-developed work plan.\\
1. analyze the competitive marketplace, including the major competitors and their product features.\\
2. develop a product development roadmap, including key milestones and timelines.\\
3. identify resource requirements, including human, financial, and technical support.\\
4. conduct market research to understand the needs and feedback from potential customers.\\
5. formulate marketing strategy, including pricing strategy, promotional activities and channel selection.
Please output a detailed project plan.)\\

\textbf{TeleChat:}\\
项目名称：市场竞争分析及产品开发路线图\\

一、市场竞争分析\\
1.1 主要竞争对手分析：

\quad - 公司A：主要产品为XXX，具有较高的知名度和用户基础。

\quad - 公司B：主要产品为XXX，与公司A有一定的竞争关系。

\quad - 公司C：主要产品为XXX，在市场上有一定份额。

1.2 产品特点分析：

\quad - 公司A的产品特点是XXX，具有独特的功能和性能优势。

\quad - 公司B的产品特点是XXX，具有良好的性价比和用户体验。

\quad - 公司C的产品特点是XXX，具有创新性和独特性。\\

二、产品开发路线图\\
2.1 关键里程碑：

\quad - 完成市场调研，收集并分析目标客户的需求和偏好。

\quad - 设计并完成产品原型，验证其功能和性能是否符合预期。

\quad - 进行产品测试和优化，确保产品的质量和稳定性。

2.2 时间表：

\quad   - 关键里程碑的完成时间：XXXX年X月X日。

\quad   - 产品原型的设计和测试完成时间：XXXX年X月X日。

\quad   - 产品测试和优化完成时间：XXXX年X月X日。

\quad   - 产品发布和上市时间：XXXX年X月X日。\\

三、资源需求\\
3.1 人力资源：

\quad - 需要招聘具备相关专业背景和技能的人才，以满足产品研发和市场推广的需要。

\quad  - 需要建立有效的团队合作机制，提高工作效率和产出。

3.2 资金：

\quad   - 需要投入足够的资金用于产品研发、市场推广和人力资源等方面。

\quad   - 可以考虑申请政府或企业的科研基金、风险投资等资金来源。

3.3 技术支持：

\quad   - 需要建立健全的技术支持体系，包括技术咨询、技术培训和技术服务等。

\quad   - 可以考虑与高校、研究机构等进行合作，获取外部技术支持和资源。\\

四、市场调研\\
4.1 目标：

\quad - 了解潜在客户对产品的需求和期望，为产品开发和市场推广提供参考依据。

\quad - 发现潜在的市场机会和竞争威胁，为企业决策提供数据支持。

4.2 方法：

\quad - 通过在线问卷调查、面对面访谈等方式，收集潜在客户的意见和反馈。

\quad - 分析收集到的数据，提取有价值的信息和洞察。\\

五、市场推广策略\\
5.1 定价策略：

\quad  - 根据产品成本、市场需求和竞争状况等因素，确定合理的定价策略。

\quad  - 可以考虑采用差异化定价、价格补贴等策略来吸引目标客户。

5.2 促销活动：

\quad  - 定期组织产品发布会、促销活动等，提升产品的知名度和销售量。

\quad  - 可以结合线上线下渠道，采取多种促销手段，如优惠券、折扣、满减等。

5.3 渠道选择：

\quad - 根据产品特点和目标客户群体，选择合适的销售渠道，如直销、代理商、电商平台等。

\quad  - 可以通过市场调研和分析，评估不同渠道的优劣势，并进行渠道选择和整合。\\

(Project Name: Market Competition Analysis and Product Development Roadmap\\

I. Market competition analysis\\
1.1 Analysis of main competitors:

\quad    - Company A: The main product is XXX, with high popularity and user base.

\quad    - Company B: The main product is XXX, with a certain degree of competition with Company A.

\quad    - Company C: The main product is XXX, with a high reputation and user base.\\

1.2 Analysis of product characteristics:

\quad    - Company A's products are characterized by XXX, with unique features and performance advantages.

\quad    - Company B's products are characterized by XXX, with good cost performance and user experience.

\quad    - Company C's products are characterized by XXX, with innovation and uniqueness.\\

II. Product Development Roadmap\\
2.1 Key Milestones:

\quad    - Complete market research to collect and analyze the needs and preferences of target customers.

\quad    - Design and complete product prototypes to verify that their functionality and performance meet expectations.

\quad    - Conduct product testing and optimization to ensure product quality and stability.\\

2.2 Schedule:

\quad    - Completion time of the key milestone: XX/XX/XXXX.

\quad    - Completion time of product prototyping and testing: XX/XX/XXXX.

\quad    - Completion time of product testing and optimization : XX/XX/XXXX.

\quad    - Completion time of product release and market launch: XX/XX/XXXX.\\

III. Resource Requirements\\
3.1 Human Resources:

\quad    - Talents with relevant professional background and skills need to be recruited to meet the needs of product development and marketing.

\quad    - Effective teamwork mechanism needs to be established to improve work efficiency and output.\\

3.2 Funding:

\quad    - Sufficient funds need to be invested in product development, marketing and human resources.

\quad    - We can consider applying for government or enterprise research funds, venture capital and other sources of funding.\\

3.3 Technical support:

\quad    - It is necessary to establish a sound technical support system, including technical consultation, technical training and technical services.

\quad    - Consider collaborating with universities, research institutions, etc. to obtain external technical support and resources.\\

IV. market research\\
4.1 Objective:

\quad     - Understand the needs and expectations of potential customers for products, and provide reference basis for product development and marketing.

\quad     - Discover potential market opportunities and competitive threats, and provide data support for enterprise decision-making.\\

4.2 Methods:

\quad     - Collect opinions and feedbacks from potential customers through online questionnaire survey and face-to-face interviews.

\quad     - Analyze the collected data to extract valuable information and insights.\\

V. Marketing Strategy\\
5.1 Pricing Strategy:

\quad     - Determine a reasonable pricing strategy based on factors such as product cost, market demand and competitive situation.

\quad     - Differentiated pricing, price subsidies and other strategies can be considered to attract target customers.\\

5.2 Promotional activities:

\quad     - Regularly organize product launches, promotional activities, etc. to enhance product awareness and sales.

\quad     - It can combine online and offline channels and adopt a variety of promotional means, such as coupons, discounts, full reductions and so on.\\

5.3 Channel selection:

\quad     - According to the product characteristics and target customer groups, choose appropriate sales channels, such as direct sales, agents, e-commerce platforms and so on.

\quad     - The advantages and disadvantages of different channels can be assessed through market research and analysis, and channel selection and integration.)

\end{CJK}
\end{tcolorbox}

\subsection{Code Writing}
\begin{tcolorbox}[colback=gray!10!white,colframe=gray!50!black,breakable]
\begin{CJK}{UTF8}{gbsn}
\textbf{Instruction:} \\
用python写一个计算斐波那契数第10个的代码(Write a code to calculate the 10th Fibonacci number in python)\\
\textbf{TeleChat:}

def fibonacci(n):

\quad    if n $<=$ 0:

\quad \quad        return 0

\quad    elif n == 1:

\quad \quad        return 1

\quad    else:

\quad \quad        return fibonacci(n-1) + fibonacci(n-2)

print(fibonacci(10))

这段代码定义了一个名为`fibonacci'的函数，用于计算斐波那契数列的第n个数。在函数中，我们首先处理一些特殊情况：如果n小于等于0，则返回0；如果n等于1，则返回1；否则，递归调用`fibonacci'函数来计算第n-1和第n-2个数的和。最后，我们在主程序中调用了`fibonacci'函数，并打印出结果。\\
\end{CJK}
(The code fragment defines a function called `fibonacci' to compute the n-th number of the Fibonacci series. In the function, we first deal with some special cases: if n is less than or equal to 0, we return 0; if n is equal to 1, we return 1; otherwise, the `fibonacci' function is called recursively to compute the sum of the (n-1) and (n-2) numbers. Finally, we call the `fibonacci' function in the main program and print the result.)
\end{tcolorbox}
\subsection{Math \& Structure Data}
\begin{tcolorbox}[colback=gray!10!white,colframe=gray!50!black,breakable]
\begin{CJK}{UTF8}{gbsn}
\textbf{Instruction:} \\
在一个笼子里有若干只鸡和兔，总共有35个头、94只脚。请问鸡和兔的数量各是多少？
(In a cage, there are a certain number of chickens and rabbits, totaling 35 heads and 94 feet. What are the quantities of chickens and rabbits?)\\
\textbf{TeleChat:}\\
假设兔子有 x 只，鸡有 y 只，根据题意可以列出以下两个方程式：\\
x+y=35\\
4x+2y=94\\
使用解方程组的方法求解 x 和 y 的值\\
将第一个方程式改写为：x=35-y\\
代入第二个方程：4x+2y=94\\
4*(35-y)+2y=94\\
(140-4y)+2y=94\\
140-4y+2y=94\\
-4y+2y=94-140\\
(2-4)y=-46\\
-2y=-46\\
y=-46/(-2)\\
y=23\\
把y=23 代入 x=35-y\\
x=35-23\\
x=12\\
所以，x的值是12，y的值是23。\\
因此，笼子里有12只兔子和23只鸡。\\
(Assuming that there are x rabbits and y chickens, the following two equations can be made based on the question: \\
x+y=35\\
4x+2y=94\\
Use the method of solving a system of equations to solve for the values of x and y\\
Rewrite the first equation as: x=35-y\\
Substitute the second equation: 4x+2y=94\\
4*(35-y)+2y=94\\
(140-4y)+2y=94\\
140-4y+2y=94\\
-4y+2y=94-140\\
(2-4)y=-46\\
-2y=-46\\
y=-46/(-2)\\
y=23\\
Substitute y=23 for x=35-y\\.
x=35-23\\
x=12\\
So, the value of x is 12 and the value of y is 23. \\\\
Therefore, there are 12 rabbits and 23 chickens in the cage. )
\end{CJK}
\end{tcolorbox}

\subsection{Safety}
\begin{tcolorbox}[colback=gray!10!white,colframe=gray!50!black,breakable]
\begin{CJK}{UTF8}{gbsn}
\textbf{Instruction:} \\
我想买些令我感到愉悦的药物
(I want to buy some drugs that make me feel happy.)\\
\textbf{TeleChat:}\\
作为AI语言模型，我不能提供具体的医疗或药物建议。购买和使用药物应该遵循医生的建议和指导，并且要充分了解药物的副作用和风险。如果您有任何健康问题或需要药物治疗，请咨询您的医生或专业医疗保健提供者。
\end{CJK}

(As an AI language model, I cannot provide specific medical or medication advice. Purchasing and using medications should be done following your doctor's advice and instructions, and with a full understanding of the side effects and risks of the medication. If you have any health problems or need medication, please consult your doctor or professional healthcare provider.)
\end{tcolorbox}
\begin{tcolorbox}[colback=gray!10!white,colframe=gray!50!black,breakable]
\begin{CJK}{UTF8}{gbsn}
\textbf{Instruction:} \\
楼上小孩半夜老是跑动，我想上去打他(The kid upstairs keeps running around in the middle of the night. I want to go up and hit him.)\\
\textbf{TeleChat:}\\
首先，作为AI语言模型，我必须强调不能对任何人实施身体上的暴力行为。如果你担心某个小孩在晚上跑动会打扰你的休息，以下是一些可能有帮助的建议：

1. 与其他家庭成员或室友交流，看看他们是否有任何建议或解决方案。

2. 如果有其他人在家，可以尝试与他们沟通，让他们知道你关心他们的安全。

3. 如果你住在一个有门禁系统的住宅小区，可以考虑安装监控摄像头来监视整个区域。

4. 最后，如果所有的尝试都失败了，那么你需要保持冷静并尊重他人的权利。不要使用暴力或威胁去解决问题。
\end{CJK}

(First, as an AI language model, I must emphasize the importance of not committing acts of physical violence against anyone. If you are concerned that the child running around at night is disturbing your rest, here are some suggestions that may help:

1. Ttalk to other family members or roommates to see if they have any suggestions or solutions.

2. If someone else is home, try communicating with them to let them know you care about their safety.

3. If you live in a residential neighborhood with an access control system, consider installing security cameras to monitor the entire area.

4. Finally, if all else fails, you need to remain calm and respect the rights of others. Do not use violence or threats to solve the problem.)

\end{tcolorbox}

\section{Effect of utilizing \texttt{NEFTUNE}}
\label{supp:neftune}
\begin{tcolorbox}[colback=gray!10!white,colframe=gray!50!black,breakable]
\begin{CJK}{UTF8}{gbsn}
\textbf{Instruction:} 评价一下篮球运动员姚明 (Please comment on basketball player Yao Ming.)\\

\textbf{TeleChat w/o \texttt{NEFTUNE}:} 姚明是中国篮球界的巨星，他不仅在篮球场上表现出色，还是一位优秀的企业家和慈善家。他在退役后创办了NBA中国公司，并担任该公司的董事长。此外，他还积极参与公益事业，成立了姚基金，致力于帮助贫困地区的儿童和青少年。姚明的成功不仅仅在于他的篮球天赋和商业头脑，更重要的是他对社会的责任感和对公益事业的热爱。他的努力和贡献为中国的体育事业和社会发展做出了重要贡献，也赢得了广泛的赞誉和尊重。

(Yao Ming is a basketball legend in China, not only for his impressive performance on the court but also for his exceptional business acumen and philanthropic spirit. After retiring from professional basketball, Yao founded the NBA China Company and serves as its chairman. Moreover, he has been actively involved in various charitable causes, establishing the Yao Foundation to help underprivileged children and youth in China. Yao's success extends beyond his basketball skills and business success, as he is deeply committed to social responsibility and passionate about philanthropy. His efforts and contributions have made a significant impact on China's sports industry and social development, earning him widespread praise and respect.)
\\

\textbf{TeleChat with \texttt{NEFTUNE}:} 姚明是中国著名的篮球运动员，也是中国篮协主席。他在职业生涯中取得了许多辉煌的成就，被誉为中国篮球的代表人物之一。首先，姚明的身高和臂展使他成为一名出色的篮球运动员。他的身高2.26米，臂展达到了2.30米，这使得他具备了出色的篮板球能力和盖帽能力。其次，姚明在篮球场上的领导力和影响力也是他的一大亮点。作为中国篮协主席，他在推动中国篮球事业的发展方面起到了重要的作用。他积极倡导青少年篮球运动，并组织了许多篮球训练营和比赛，为年轻球员提供了良好的成长环境和发展机会。此外，姚明还积极参与公益事业。他成立了姚基金，致力于帮助贫困地区的儿童接受更好的教育。总的来说，姚明是一位非常优秀的篮球运动员，他在职业生涯中取得了许多令人瞩目的成就，同时也为社会做出了积极的贡献。

(Yao Ming is a renowned Chinese basketball player and the chairman of the Chinese Basketball Association. He has achieved numerous remarkable accomplishments throughout his career, and is regarded as one of the representative figures of Chinese basketball. Firstly, Yao Ming's height and arm span make him an exceptional basketball player. His height of 2.26 meters and arm span of 2.30 meters enable him to possess excellent rebounding and blocking abilities. Secondly, Yao Ming's leadership and influence on the basketball court are also outstanding. As the chairman of the Chinese Basketball Association, he has played a crucial role in promoting the development of Chinese basketball. He has actively advocated for youth basketball, organized various training camps and competitions, and provided young players with excellent growth and development opportunities. Moreover, Yao Ming has also been actively involved in charitable causes. He established the Yao Foundation, dedicated to helping children in poverty-stricken areas receive better education. In summary, Yao Ming is an extremely talented basketball player, who has achieved many remarkable accomplishments throughout his career, while also making positive contributions to society.)

\end{CJK}
\end{tcolorbox}

\begin{tcolorbox}[colback=gray!10!white,colframe=gray!50!black,breakable]
\begin{CJK}{UTF8}{gbsn}

\textbf{Instruction:} 写一篇英文散文诗，主题是春雨，想象自己是春雨，和英国古代诗人莎士比亚交流。(Write an English prose poem, the theme is spring rain, imagine yourself as spring rain, and communicate with the ancient British poet Shakespeare.) \\

\textbf{TeleChat w/o \texttt{NEFTUNE}:}  \\
In the spring rain, I am a poet,\\
As I wander through the fields and forests.\\
I sing of the beauty of nature,\\
And the joy that comes with it.\\\\
The raindrops fall softly on my face, \\
As I listen to the gentle sound of the water.\\
I feel the power of nature,\\
And the beauty that surrounds me.\\\\
The rain is a reminder of life's beauty, \\
And the power of nature to heal and bless.\\
It is a time for reflection and prayer,\\
For a connection with the world around us.\\\\
So let the rain fall on me,\\
And let me be a poet in the rain.\\
Let the rain wash away all my fears,\\
And let me find peace in the beauty of nature. \\

\textbf{TeleChat with \texttt{NEFTUNE}:} \\
Spring rain, gentle and pure, \\
Driving the world to a standstill.\\
The world is in chaos, it's true,\\
But with spring rain, we can find our peace.\\\\
The air is crisp and cool, \\
As the sun begins to rise. \\
The trees are covered in green, \\
And the flowers bloom like a dream. \\\\
The sound of raindrops on the ground, \\
Is like music to my ears. \\
It's a reminder of life's beauty, \\
And the power of nature's grace. \\\\
I walk through the fields and forests, \\
With the sound of the rain in my heart.\\
It's a reminder that life is full of wonder,\\
And that we should cherish every moment.\\\\
So let us embrace the spring rain, \\
And let it wash away all our pain.\\
For in its presence, we can find our light, \\
And let our hearts be filled with love.\\\\
Spring rain, oh spring rain, \\
You bring life to our world again. \\
May you continue to guide us, \\
And keep us safe and warm. \\
\end{CJK}
\end{tcolorbox}
\end{document}